\documentclass[letterpaper, 10 pt, conference]{ieeeconf}  % Comment this line out
\IEEEoverridecommandlockouts                              % This command is only
\usepackage[utf8]{inputenc}
\usepackage[T1]{fontenc}
\usepackage[backend=biber,style=ieee,sortlocale=de_DE,natbib=true,url=true,doi=false,eprint=false]{biblatex} %issn= false
\usepackage[disable]{todonotes}
\usepackage[destlabel=true,colorlinks=true,urlcolor=blue,linkcolor=blue,citecolor=blue]{hyperref}
\usepackage{soul}
\usepackage[]{float} %% caption= false
\usepackage{mathtools}
\usepackage{booktabs}

\usepackage{tikz}
\usepackage{amsfonts}

\newtheorem{definition}{Definition}

\usepackage{listings}
\usetikzlibrary{arrows.meta}
\usepackage{hyperref}

\usepackage{lipsum}
\usepackage{algorithm}
\usepackage[noend]{algpseudocode}

\usepackage{pgfplots}
\usepgfplotslibrary{groupplots}
\pgfplotsset{compat=1.18}
\usetikzlibrary{patterns,shapes.arrows}

\newcommand{\va}{\mathbf a}

\newcommand{\vf}{\mathbf f}

\newcommand{\vh}{\mathbf h}

\newcommand{\vm}{\mathbf m}

\newcommand{\vq}{\mathbf q}

\newcommand{\vu}{\mathbf u}
\newcommand{\vv}{\mathbf v}

\newcommand{\vx}{\mathbf x}

\newcommand{\vP}{\mathbf{P}}
\newcommand{\vQ}{\mathbf{Q}}
\newcommand{\vR}{\mathbf{R}}

\usepackage{subcaption}
\usepackage{placeins}

\usetikzlibrary{arrows}
\usetikzlibrary{shapes}

\title{Automatic Configuration of Multi-Agent Model Predictive Controllers\\ based on Semantic Graph World Models}

\author{K. de Vos$^{1,*}$, E. Torta$^{1}$, H. Bruyninckx$^{1,2,3}$, C.A. López Martínez$^{1}$, M.J.G. van de Molengraft$^{1}$ 
\thanks{$^{1}$ Department of Mechanical Engineering, Eindhoven University of Technology, The Netherlands}%
\thanks{$^{2}$Department of Mechanical Engineering, KU Leuven, Belgium}
\thanks{$^{3}$Flanders Make, Leuven, Belgium}
\thanks{$^{*}$E-Mail: {\tt\small k.d.vos at tue.nl}}
}

\begin{document}

\maketitle
\thispagestyle{empty}
\pagestyle{empty}

\begin{abstract}
We propose a shared semantic map architecture to construct and configure Model Predictive Controllers (MPC) dynamically, that solve navigation problems for multiple robotic agents sharing parts of the same environment.
The navigation task is represented as a sequence of semantically labeled areas in the map, that must be traversed sequentially, i.e. a route. Each semantic label represents one or more constraints on the robots' motion behaviour in that area. The advantages of this approach are: (i) an MPC-based motion controller in each individual robot can be (re-)configured, at runtime, with the locally and temporally relevant parameters; (ii) the application can influence, also at runtime, the navigation behaviour of the robots, just by adapting the semantic labels; and (iii) the robots can reason about their need for coordination, through analyzing over which horizon in time and space their routes overlap. The paper provides simulations of various representative situations, showing that the approach of runtime configuration of the MPC drastically decreases computation time, while retaining task execution performance similar to an approach in which each robot always includes all other robots in its MPC computations.
\end{abstract}

%% input sections/chapters which are located within the chapters folder
\section{Introduction}

% The .. of robotic systems in applications such as warehousing and automated farming.  

% The use of autonomous mobile robots in applications such as warehousing and automated farming is becoming increasingly common. In the former application there is a need for high throughput, the latter demands robots that are highly specialised for a certain task. Both, therefore, have a need for multi-robot systems to fulfil those requirements. Apart from performance, a key aspect of the design of these systems is safety, especially the avoidance of collisions with each other, elements of the environment, or humans. 

Multi-robot deployment has become common in applications such as warehousing, manufacturing, and farming. This requires motion coordination approaches that guarantee  safe navigation of all robotic agents that share the same, often dynamic, environment. These environment typically contain an inherent structure, such as the aisles and intersections in a warehouse setting. The environment dynamics only occur through obstacles that move through the environment.

% For large scale multi-robot systems, Model Predictive Controllers (MPC) provides a balance of reactivity and planning. However, it also presents challenge with respect to its configuration. Especially when considering homogeneous robots, dynamic environments, and varying task specifications. 
In this work we present a method to dynamically configure Model Predictive Controllers to solve local navigation problems based on the information represented in the semantic graph world model.
Model Predictive Control (MPC) is a motion control approach that solves a constrained optimisation problem at every sample time, and applies the control input in a receding horizon fashion. Since constraints and objectives of agents can be composed independently of each other, MPC is applicable for solving the navigation coordination challenges in the above-mentioned applications, in which agents and their task descriptions are expected to be heterogeneous, and environments dynamic. 

% that fits very well to the navigation coordination challenges in the above-mentioned applications: as long as every individual robot can take into account the constraints that are introduced by the environment, the routes, the other robots, and the obstacles, it can optimize its own motion independently of the controllers in the other robots. This is an especially attractive approach when the robots have heterogeneous motion and perception capabilities, the environment is dynamic, as are the individual robots' task specifications.

% In this paper we propose a method that can automatically derive the constraints and objectives of an MPC from a graph-world model which describes the layout and geometry of the environment, which is assumed to be known and structured.
% Within the scope of this paper structured environments are environments that possess some inherent structure, e.g. aisles in a warehousing environment or roads in a conventional traffic setting. We do, however, not assume that this structure is undisturbed by, for instance, an unmapped or dynamic obstacle. Within such an environment a team of robotic agents is deployed. Each robot is tasked with navigating from its starting location, via a set of intermediate areas, to a final location. While executing their task the individual robots must coordinate their motions such that collisions with elements of the environment and each other are avoided.

We show that constraints and objectives of the MPC problem can be derived from a semantic graph world model that represents the environment's layout, geometry, and area semantics. We argue that by changing the area semantics, the layout or the geometry of the environment as represented in the graph world model, the robot will behave differently in different areas in the environment for several reasons, such as closeness to walls, presence of obstacles, constraints on maximal or minimal speed, etc.
The navigation task of a robot (its route), is represented as a sequence of areas it has to drive through. While executing their task the MPC controller ensures that individual robots coordinate their motions such that collisions with elements of the environment and each other are avoided.
As reported in e.g.~\cite{Parker2008}, communication can significantly boost the coordination performance, especially in complex dynamic environments. However, communication is not always necessary, since the risk of collision varies as robots get closer together or further away from each other. More importantly, when the number of agents becomes large, the broadcasting of communication messages between all agents in an environment may become infeasible.  

%\todo[inline]{Solving centrally reduces communication, but is infeasible when the number of agents grows large. "This is computationally expensive when M is large, making
%the centralized formulation not scalable with the number
%of robots" \cite{firoozi2020distributed} }

%\todo[inline]{glue}

Our contribution is twofold.
Firstly we propose a semantic graph world model that captures both the topology and the geometry of the environment in the form of a property graph. The topology of the graph is designed to allow agents to dynamically and automatically derive which constraints and objective functions are applicable to their MPC controller given the current state of the system as represented in the graph.  Secondly, we propose an algorithm to dynamically determine which agents should coordinate their movements which results in a reduction of the number of constraints in the MPC controller of each agent with a significant reduction in communication and computation time of the MPC controller. 

%% Structure of the paper needs to go here

%% add 
 %The advantages of this approach are ....
 %
 %\todo[inline]{Cesars input: Integrated planning and control. Taking into account kinematic constraints at all level. Smoother control actions. Planning at necessary resolution. Coarse planning, more compatible with topological planning, more scalable. Included above not happy yet.}

\textit{Related work.}
%\cite{doi:10.5772/57313}
Different methods can be used to solve multi-agent navigation problems. Planning is often performed offline assuming static or predictable system's  configurations (see  \cite{standley2011complete,WAGNER20151, standley2011complete, 7138650} for examples). When deviations between the plan and the system's state are observed, a new plan can be computed and executed. Contrary to planning, reactive approaches (see \cite{580977, fiorini1998motion, 4543489} for examples) tend to adapt to rapid changes in the environment at the expense of disregarding global optimality, deadlock or livelock freeness. 
A combination of planning and reactivity is beneficial for reliable navigation in multi-robot systems. Model Predictive Control (MPC) is a method that combines planning and predictions over a defined horizon with reactivity to the current system's state therefore it is regarded as a viable method to address multi-robot navigation problems.
In MPC-based approaches, trajectories are computed by solving a constrained optimization problem. The objective function is commonly a measure of the desired performance of the system. Constraints typically arise from the system dynamics, hardware limits, and (safety-) requirements. Within the context of mobile-robot navigation, different MPC formulations have been proposed, e.g.,  \cite{mercy2018online, ferranti2018coordination}. Similarly to the scenarios addressed by this work,  \cite{mercy2018online} consider robot navigation in environments with an inherent structure. They introduce the use of separating hyperplane constraints for collision avoidance which this paper adopts in the proposed MPC formulation as well.
On the same line but for vessels navigation, \cite{ferranti2018coordination}, designed a distributed nonlinear MPC for the navigation of vessels at a canal intersection. 
%
%In \cite{9981780} a framework is described to tune the MPC controller of an autonomous racecar based on the current environmental conditions using a Bayesian optimisation approach.  

MPC approaches, however, often suffer from issues with scalability as the number of agents grows large \cite{firoozi2020distributed}. To address scalability issues, many authors have proposed decentralizing the MPC problem \cite{ferranti2018coordination,serra2020whom, mercy2018online}. This, however, comes at the cost of an increased need for communication due to the coupling between sub-problems resulting from agents sharing the same workspace.
Several works have proposed reducing the complexity of the MPC problem itself. The proposed methods typically use the current context to reduce the number of optimisation variables and/or constraints, or to decouple problems which are unlikely to influence each other. The authors of \cite{serra2020whom} propose a framework in which a multi-agent MPC trajectory planner is deployed alongside a communication policy that, using a multi-agent reinforcement learning approach, has learned when and with whom the agent should communicate to avoid collisions. 
%The resulting agents show similar behavior to a broadcasting approach, while significantly reducing communication needs. They, furthermore, show improved performance over distance-based heuristic approaches. 
The authors of \cite{mercy2018online} combine a global path planner with a local trajectory generator. This approach allows them to divide the task of navigating a large and complex environment into smaller frames, reducing the scope and complexity of subsequent optimal control problems. This is in line with the method presented in our paper however, the division in smaller frames is here achieved by interpreting the system state as provided by a graph world model.
Prior work has explored the use of graph-based semantic world models to automatically generate semantic maps for localization \cite{Hendrikx} or navigation \cite{PAUWELS2023101959}, in this paper we propose a new method to automatically (re)configure the (MPC-based) controller itself by automatically determining the applicable constraints and objective function.

\section{Method}

\subsection{MPC Formulation}
% i,j agent
% k time step
% l agent mode
% m element counter 
% t absolute time
%Model Predictive Control (MPC) is an optimisation method that can be used to compute optimized trajectories of agents considering their current configuration, their dynamics, and the constraints which they are subjected to. 
% Within MPC the predicted motion of agents is computed over an horizon N. The latter is especially of interest in multi-agent system, in which the trajectories of individual agents can potentially intersect. If intersections can be predicted they can be acted upon before they occur.\\
We cast the solution of the local navigation problem as an MPC problem. Since, within the MPC framework, collisions between the intended motion trajectories of individual agents can be predicted and acted upon before they occur. We consider an environment in which a set of agents $\mathbb{A}$ is deployed with $\mathbb{F}_o \subseteq \mathbb{A}$  a subset of agents who should coordinate their movements. Given a set of Objectives $\mathbb{O}_i$ and a set of environment elements, such as walls and obstacles, $\mathbb{W}_i$ for each individual agent $A_i \in \mathbb{F}_o$ the multi-agent MPC problem can be formulated as:
\allowdisplaybreaks[4]
\begin{align}
    &\min_{\vx, \vu} \sum_{i = 0}^{|\mathbb{F}_o|} J_i(\vx_i, \vu_i, m_i, t),
\end{align}

Subject to:
\begin{align}
    \quad  &\vx_i(t + k + 1) = \vf_i (\vx_i(t + k), \vu_i(t + k)),\\
    &\vx_i(t) = \vx_{i, init},\\
    & \vx_{i,min} \leq \vx_i(t + k) \leq \vx_{i,max},\\
    & \vu_{i,min} \leq \vu_i(t + k) \leq \vu_{i,max},\\
    &\text{dist}\left(A_i\left(t + k\right), w\right) > 0, \quad  \forall w \in \mathbb{W}_i, \\
    &\text{dist}\left(A_i\left(t + k\right), A_j\left(t + k\right)\right) > 0,\quad \forall A_j \in \mathbb{F}_j \land j\neq i,
\end{align}
in which $\vf_i$ describes the dynamic model of agent $i$, which within the context of this paper is assumed to reflect a kinematic unicycle model. Furthermore, $\vx_i \in \mathbb{X}_i$ and $\vu_i \in \mathbb{U}_i$ represent the state and input of agent $i$ respectively. The state of agent $i$ describes its longitudinal velocity and its Cartesian position and orientation w.r.t a global reference frame. The control inputs of agent $i$ are the acceleration in longitudinal direction and its angular velocity. 
%Conceptually, these inputs correspond to, respectively, changing the gas pedal position and steering angles in a car; so, ``sampling times'' of the MPC computations of the order of 1 second are often fast enough.
%\todo[inline]{Overlap in indices between $\mathbb{F}_i$ and $A_i$}
\subsubsection*{Objective}

%Within the MPC formulation 
We employ a standard cost function structure, containing intermediate costs $l_i$ and final costs $n_i$,
\begin{align}
    J_i(\vx_i, \vu_i, m_i, t) := \sum_{k = 0}^{N_t} &l_i(\vx_i(t + k), \vu_i(t+k))\\
    &+ n_i(\vx_i(t + k)),
\end{align}
with $t$ the current time step, and $\vx_i \in \mathbb{X}_i$ and $\vu_i \in \mathbb{U}_i$ the predicted state-trajectory and optimised input-trajectory of agent i respectively computed over a horizon of length~$N_t \in \mathbb{N}$.
%Within the context of this paper 
The intermediate cost function is defined~as
\begin{equation}
    l_i(\vx_i,\vu_i,t) := \vu_i^T \vR_i \vu_i,
\end{equation}
with $\vR_i$ a cost matrix of appropriate dimension that weights the system input of agent $i$. The terminal cost function rewards progress made within the prediction horizon and is thus defined as
\begin{align}
    n_i(\vx_i, t, m_i) &:= \sum_{j=0}^{|\mathbb{O}_i|} w_o \left[ \ d_{i,j}^T \ \vQ_{i,j} \ d_{i,j} + \vq_{i,j}^T d_{i,j}, \right] %\\
             %&\phantom{:}= \vx^T \vA^T \vQ \vA \vx + 2 \vB^T \vQ \vA x + \vB^T \vQ \vB\
\label{eq:costfun}
\end{align}
with $d_{i,j}$ the (signed) distance of agent $A_i$ from the $j$-th element of the set of objectives $\mathbb{O}_i$, with $w_o$ weighing the objectives. The weights are determined through analysis of the solution in the previous iteration. All objectives corresponding to areas which were reached in the previous prediction, are weighted equally, the remaining objectives receive zero weight.
%\todo[inline]{Do we need to be more specific on the determination of the weights. They are determined based on the mode of the end of the prediction horizon}
% \begin{equation}
%     d_i(\vx, t):=  \vA_i(l_i) \vx_i(t) + \vb(l_i)
% \end{equation}

% \begin{align}
%     \vA_i(l_i) &:=
%     \begin{bmatrix}
%         a^{*}, b^{*}
%     \end{bmatrix},&
%     \vb_i(l_i) &:= c^*
% \end{align}

% \begin{equation}
%     p^* := \frac{p}{\sqrt{a^2 + b^2}}
% \end{equation}

\subsubsection*{Constraints}
For safety reasons, agents should not collide with any elements, e.g. walls, obstacles or virtually defined boundaries, in the environment, nor with each other. To represent the no-collision constraints we utilize the separating hyperplane constraint \cite{OMGtools2016}. This allows the specification of no-collision constraints between agents and obstacles described by convex shapes, however, puts no limitation on the convexity of the resulting drivable space. This opposed to, for example, formulating the non-collision constraints as a collection of linear constraints which only allows for describing a convex drivable space.

For each element $w \in \mathbb{W}_i$ with vertices $\vv_{w}$, we formulate the no-collision constraints for a circular~agent $A_i$ with radius~$r_v$~as:
\begin{align}
    \va_{w,i}^T \vv_{w,j}- b_{w,i} &\geq 0,   &&\forall \vv_{w,j} \in \vv_w,\\
    \va_{w,i}^T \vx(k) - b_{o,i} &\leq -r_v,  &&\forall k \in [0,N_t],\\
    ||\va_{w,i}||_2 &\leq 1,
\end{align}
with $\va$ and $b$ the normal vector to and offset of the hyperplane respectively. Since we approximate the robot's geometric footprint with a circle, the no-collision constraint between any two agents $i$ and $j$, with radii $r_i$ and $r_j$ respectively, can be formulated as
% \begin{figure}[t]
%     \centering
%     \resizebox{0.4\linewidth}{!}{\input{Figures/hyperplane}}
%     \caption{Example solution of the hyperplane constraints for two different agents, with radius $r_v$, with two distinct rectangular obstacles.}
% \end{figure}
\begin{align}
    || \vx_{k,i} - \vx_{k,j} ||_2 &> (r_i + r_j), &\forall k &\in [0, N_t].
\end{align}
To encourage trajectories further away from obstacles and other agents, all no-collision constraints are augmented with soft constraints with an enlarged agent radius of $r_{soft}$.

\subsection{Semantic Map}
\begin{figure}[t]
    \centering
    \begin{subfigure}[b]{0.335\linewidth}
    \resizebox{\linewidth}{!}{\begin{tikzpicture}[align=center,node distance=1.5cm]
    %% Spaces
    \draw  [dashed] (-3.5,-2) node (v6) {} rectangle (-2.5,2) node (v2) {} node[pos=0.5] {$S_0$};
    \draw  [dashed] (-3.5,2) node (v1) {} rectangle (-2.5,3) node (v3) {} node[pos=0.5] {$S_1$};
    \draw  [dashed] (-2.5,3) rectangle (0.5,2) node (v5) {} node[pos=0.5] {$S_2$};

    %% Walls
    \draw  [fill = {rgb:black,1; white,10}, draw = black] (-3.55,-2) rectangle (-3.75,3.25);
    \draw  [fill = {rgb:black,1; white,10}, draw = black] (-3.55,3.25) rectangle (0.5,3.05) node (v4) {};
    \draw  [fill = {rgb:black,1; white,10}, draw = black] (-2.45 , -2) node (v7) {} rectangle (-2.25,1.95);
    \draw  [fill = {rgb:black,1; white,10}, draw = black] (0.5, 1.75) rectangle (-2.45, 1.95);

    \node (v8) at (0.5,3) {};
    \draw  [red] (v6.center) -- (v7.center) node [midway, above] {$I_0$};
    \draw  [red] (v1.center) -- (v2.center) node [midway, below] {$I_1$};
    \draw  [red] (v2.center) -- (v3.center) node [midway, right] {$I_2$};
    \draw  [red] (v5.center) -- (v8.center) node [midway, left] {$I_3$};
    
\node at (-1,1.5) {$W_3$};
\node (v9) at (-4.2,0) {$W_0$};
\node at (-1,3.5) {$W_2$};
\node at (-1.8,0) {$W_1$};
\end{tikzpicture}}
    \caption{}
    \label{fig:semi_struc_env}
    \end{subfigure}
    ~
    \begin{subfigure}[b]{0.35\linewidth}
    \resizebox{\linewidth}{!}{\usetikzlibrary{positioning}

\usetikzlibrary{arrows}
\begin{tikzpicture}[align=center,node distance=1.75cm]

\definecolor{myred}{rgb}{0.906,0.298,0.235}
\definecolor{myblue}{rgb}{0.204, 0.596, 0.859}
\definecolor{myyellow}{rgb}{0.992,0.796,0.431}
\definecolor{mygray}{rgb}{0.698, 0.745, 0.765}
\definecolor{mygreen}{rgb}{0.152, 0.682, 0.376}
\definecolor{mypurple}{rgb}{0.698, 0.745, 0.765}
\definecolor{mybrown}{rgb}{0.882, 0.439, 0.333}

%% Space 1
\node [circle, draw] (S1) at (0, 0) {\normalsize $S_1$};
\node [circle, draw, below = of S1, fill = mygreen] (I1) {\normalsize $I_1$};
\node [circle, draw, right = of S1, fill = mygreen] (I2) {\normalsize $I_2$};

% Interface 1
\node [circle, draw, right = of I1, fill = mybrown] (O0) {\normalsize $O_0$};
\node [circle, draw, left = of I1, fill = myblue] (E0) {\normalsize $E_0$};

% Interface 2
\node [circle, draw,  above = of I2, fill = myblue] (E1) {\normalsize $E_1$};
\node [circle, draw,  left = of E1,  fill = mybrown] (O1)  {\normalsize $O_1$};

% Space 0
\node [circle, draw, below = of I1] (S0) {\normalsize $S_0$};
\node [circle, draw, left = of S0, fill = myyellow] (W0) {\normalsize $W_0$};
\node [circle, draw, right = of S0, fill = myyellow] (W1){\normalsize $W_1$};

% Space 2
\node [circle, draw, right = of I2] (S2) {\normalsize $S_2$};
\node [circle, draw, below = of S2, fill = myyellow] (W3) {\normalsize $W_3$};
\node [circle, draw, above = of S2, fill = myyellow] (W2) {\normalsize $W_2$};

\draw [-latex]  (S1) edge node [sloped, anchor = center, above] { \small hasInterface} (I2);
\draw [-latex]  (S2) edge node [sloped, anchor = center, above] { \small hasInterface} (I2);
\draw [-latex]  (S0) edge node [sloped, anchor = center, above] { \small hasInterface} (I1);
\draw [-latex]  (S1) edge node [sloped, anchor = center, above] { \small hasInterface} (I1);

\draw [-latex]  (S2) edge node [sloped, anchor = center, above] { \small hasBoundary} (W3);
\draw [-latex]  (S2) edge node [sloped, anchor = center, above] { \small hasBoundary} (W2);

\draw [-latex]  (S0) edge node [sloped, anchor = center, above] { \small hasBoundary} (W0);
\draw [-latex]  (S0) edge node [sloped, anchor = center, above] { \small hasBoundary} (W1);

\draw [-latex]  (I1) edge node [sloped, anchor = center, above] { \small hasEvent} (E0);
\draw [-latex]  (I1) edge node [sloped, anchor = center, above] { \small hasObjective} (O0);

\draw [-latex]  (I2) edge node [sloped, anchor = center, above] { \small hasObjective} (O1);
\draw [-latex]  (I2) edge node [sloped, anchor = center, above] { \small hasEvent} (E1);

\end{tikzpicture}}
    \caption{}
    \end{subfigure}
    \caption{(Left) Simplified structured environment consisting of three areas. (Right) Excerpt of a graph world model describing the environment on the left. The colors of the nodes signify their different types.}
    \label{fig:semi_struc_graph}
\end{figure}
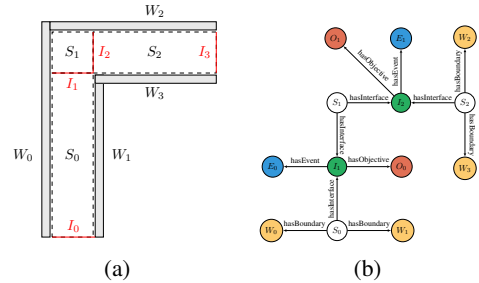

Similarly to \cite{Hendrikx,PAUWELS2023101959}, the semantic map is a graph world model that describes the environment in which the agent is deployed at a topological level: 
\noindent
\begin{definition}[Semantic Map]\label{def:semmap}
a graph $G=(V,E)$, with vertices $V$ of type \{Area, Boundary, Interface\}, and edges $E$ between vertices that show a spacial connection.
\end{definition}
Knowledge of the environmental geometry is represented as a property of the vertices composing the graph.
We assume each agent has access to an identical world model which contains correct information on the current state of the environment. We also assume that agents have the ability to recognize new or moved obstacles and are able to update the graph accordingly.
A simple example environment, with corresponding graph world model, is depicted in Figure~\ref{fig:semi_struc_graph}. The vertices of the graph represent semantic elements which we refer to as semantic map primitives. Three semantic primitives are identified: Areas, Interfaces, and Boundaries.
\textbf{Areas} represent the drivable space for the agent, and are, therefore, the building blocks of the route of each agent. \textbf{Interfaces} model the connectivity between areas, i.e., areas are not directly connected through an edge in the graph but always through one interface node between them. Within the current area, interfaces are either active or inactive. 
Active interfaces are only those interfaces which connect the current area with the next one according to the route. %
Inactive interfaces are thus those interfaces that would lead the agent, if crossed, to areas that are not part of the current route. 
Crossing inactive interfaces should thus be prevented, through adding them as constraints in the local navigation problem. The navigation through an area is, furthermore, constrained through the addition of \textbf{Boundaries}. These can stem from physical elements such as walls or obstacles or can be defined virtually to prevent the agents from entering certain regions such as the space in front of emergency doors. For safety reasons, these are interpreted as hard constraints. %Within the context of this paper, we define the semantic map containing these primitives, as follows:
We furthermore specify properties of semantic map primitives. These properties can stem from the geometry of those elements, which are encoded as lines and polygons, or can add additional information for the construction of the MPC problem. More specifically, interfaces provide information on the objective, events, and virtual boundaries. We define a mapping between the nodes, edges and properties of the graph and the configuration parameters of the MPC controller. In particular, the line geometry of nodes of type interface are mapped to the objective in \eqref{eq:costfun}, virtual boundaries are mapped to non-collision constraints to prevent the agent from entering areas outside of its route, events encode the transitions between areas, and are monitored to detect the agents entering the next area in the sequence, thus triggering a reconfiguration of the MPC controller.

\subsection{Navigation Strategy}

% Commonly used navigation approaches often specify the task description of a navigation task either as tracking a pre-specified path or trajectory, or as reaching a certain final pose \cite{}. In this work we take a different approach.
We assume agents are already assigned the route, i.e., the sequence of areas to be traversed sequentially. The navigation task is thus decomposed into sub-tasks. Each sub-task requires the agent to reach the end of the area by reaching the corresponding interface, while avoiding all collisions with other agents or environmental elements.
For example, consider again the environment depicted in Figure~\ref{fig:semi_struc_graph}, and an agent located in Area $S_0$. Instead of formulating the navigation task as reaching a certain coordinate, the navigation task can be formulated by specifying a sequence of areas to be visited, in this case, the task specification is \lq\lq\textit{reach $I_3$ after crossing areas $S_0$, $S_1$ and $S_2$}\rq\rq. From their routes, the agents can dynamically configure their MPC controller.

Specifying the local navigation tasks as a set of subsequent sub-tasks, combined with the semantic map in Definition \ref{def:semmap}, allows the agent to reason about which elements in the world are relevant in the current area and for its current sub-task, e.g. walls or agents which are only connected to topologically far-away areas are not relevant at this time instance, which reduces the complexity of the navigation problem. This thus enables the simplification of the local navigation problems.

\begin{figure}[t]
    \centering
    \resizebox{0.45 \linewidth}{!}{\input{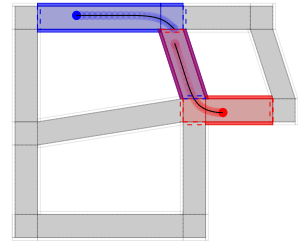}}
    \caption{Illustration of two agents with overlapping semantic horizons with $N_{h} = 2$. Red and blue polygons illustrate the environmental no-collision constraints of the red and blue agent respectively. Purple polygons show the overlap between the two. Dashed lines illustrate the virtual boundary constraints. }
    \label{fig:semhorizon}
\end{figure}

In other words, given the semantic map in Definition \ref{def:semmap}, the multi-agent navigation problem can be specified by considering a team of $N_A$ agents that are deployed in an environment of which a graph world model that follows Definition \ref{def:semmap} is available.
Agent $A_i$ is tasked with moving from its current position $\vx_{i,0}$ in area $S_{i,0}$ to its goal area $S_{i,g}$. A planned sequence of sub-tasks $\vP_i = [S_{i,0} \dots S_{i,g}]$ to complete its task is provided to the agent, e.g. by a fleet-management system. It is assumed that, under nominal undisturbed conditions, execution of plan $\vP_i$ leads to a successful completion of the task, i.e. that both task and plan are feasible. The task is deemed to be successfully completed once agent $A_i$ is contained within area $S_{i,g}$ without having collided with any other agent $A_j$ or with any element of the environment. Within our approach, we define that the index $m_i$ of an agent $A_i$ corresponds to the index of the area in the plan that is currently traversed by the agent.
The objective of agent $A_i$ is the objective encoded by the interface that connects its current area $S_{i}$ with the next area of its route. The set of hard environmental constraints $\mathbb{W}_i$ to which the agent $A_i$ is subjected contains only those (virtual-) boundaries which are linked via vertices to its current area  $S_{i}$. The inter-agents constraints are formulated between couples of agents $c_i := (A_i, A_j)$ which are contained in the same area. Once the agent reaches the currently active interface, and its event is triggered, the agent has entered the area succeeding its current area, the objective and constraints are updated accordingly. 
To decrease the influence of the discrete transitions in objective and constraints between the current and the next area, a semantic horizon $\vh$ is introduced, which acts as a look-ahead on the objectives and constraints of the next $N_{h}$ areas in the pre-specified planning. It effectively means, that not only does the agent take into account the constraints and objectives in the current area $S_{i,m_i}$, but it, furthermore, takes into account the environmental elements and objectives which relate to the next $N_h$ areas of the pre-specified route. The sets of coordinating agents, and thus the set of agents controlled by each MPC controller, are determined by checking the overlaps between their semantic horizon with length $N_{hA}$. Agents will cooperate if there is a common area in this semantic horizon. The problems from sets of agents which do not show overlap in their horizons can be solved separately. Figure \ref{fig:semhorizon} depicts the semantic horizon, and corresponding constraints, for an example scenario. The algorithm for retrieving relevant elements and the algorithm for retrieving the relevant agent configuration, given the graph world model $K$, are reported on in more detail in algorithms \ref{alg:rel_el} and \ref{alg:rel_ag} respectively, these algorithms are called at startup, and thereafter every time one of the agents leaves its current area and enters the next.

\begin{algorithm}[t]
    \caption{Relevant element retrieval for agent $A_i$}\label{alg:rel_el}
    \begin{algorithmic}
    \State \textbf{Input:} $K$, $\mathbb{P}_i$, $m_i$, $N_{h}$
    \For{ $j \gets m_i$ to $m_i + N_{h}$  }
        \State $\mathbb{B}_{i,j} \gets \text{Boundaries from } $K$ \text{ related to } \mathbb{P}_i [j] $
        \State $\mathbb{V}_{i,j} \gets \text{Interfaces from } $K$ \text{ related to } \mathbb{P}_i [j] \land \neg 
        \mathbb{P}_i [j \pm 1]$
        \State $\mathbb{W}_{i,j} \gets \mathbb{B}_{i,j}  \cup \mathbb{V}_{i,j} $
        \State $\mathbb{I}_{i,j} \gets \text{Interfaces from } $K$ \text{ related to } \mathbb{P}_i [j] \land \mathbb{P}_i [j + 1]$
        \State $\mathbb{O}_{i,j} \gets \text{Objectives from } $K$ \text{  related to } \mathbb{I}_{i,j}$
        \State $\mathbb{E}_{i,j} \gets \text{Events from } $K$ \text{ related to } \mathbb{I}_{i,j}$
    \EndFor
    \State \textbf{Return:} $\mathbb{W}_i$, $\mathbb{I}_i$, $\mathbb{O}_i$, $\mathbb{E}_i$
    \end{algorithmic}
\end{algorithm}
\begin{algorithm}[t]
    \caption{Relevant agent configuration retrieval}\label{alg:rel_ag}
    \begin{algorithmic}
        \State \textbf{Input:} $K$, $\mathbb{A}$, $\mathbb{P}$, $\vm$, $N_{sh}$
        \State $c \gets \emptyset$, \hspace{0.25 em} $\mathbb{A}_f \gets \emptyset$
        \State \textbf{\textit{Find couples of agents which have overlap}}
        \For{ $i \gets 0$ to $N_{A}$}
            \State $\vh_i \gets \mathbb{P}_i[m_i:m_i + N_{hA}]$
            \For{$j \gets i+1$ to $N_{A}$}
                \State $\vh_j \gets \mathbb{P}_j[m_j:m_j + N_{hA}]$
                \If{$\vh_i \cap \vh_j \neq \emptyset$}
                    \State $c \gets c \cup \{(A_i, A_j)\}$
                    \State $\mathbb{A}_f \gets \mathbb{A}_f \cup \{A_i\} \cup \{A_j\}$
                \EndIf
            \EndFor
        \EndFor
        \State \textbf{\textit{Expand sets of relevant agents}}

        \State $\mathbb{F}^k  \gets c$ \Comment{Sets at iteration 0}
        \While{$ \mathbb{F}^k \neq \mathbb{F}^{k+1} $}
        \State $\mathbb{F}^{k+1} \gets \emptyset$
        \For{$\mathbb{F}_i \text{ in } \mathbb{F}^k$}
            \For{$\mathbb{F}_j \text{ in } \mathbb{F}^{k+1}$}
                \If{ $\mathbb{F}_i \cap \mathbb{F}_j \neq \emptyset$ }
                    \State $\mathbb{F}_i \gets \mathbb{F}_i \cup \mathbb{F}_j$
                \EndIf
            \EndFor
            \State $\mathbb{F}^{k+1} \gets \mathbb{F}^{k+1} \cup \{\mathbb{F}_i\}$
        \EndFor
        \EndWhile
        \State \textbf{\textit{Add agents without overlap}}
        \State $\mathbb{F} \gets \mathbb{F} \cup ( \mathbb{A} \setminus \mathbb{A}_f)$
        \State \textbf{Return:} $\mathbb{F}$ \Comment{Set of sets of relevant agents}
        \end{algorithmic}
\end{algorithm}

\section{Validation}

\begin{figure*}[t]
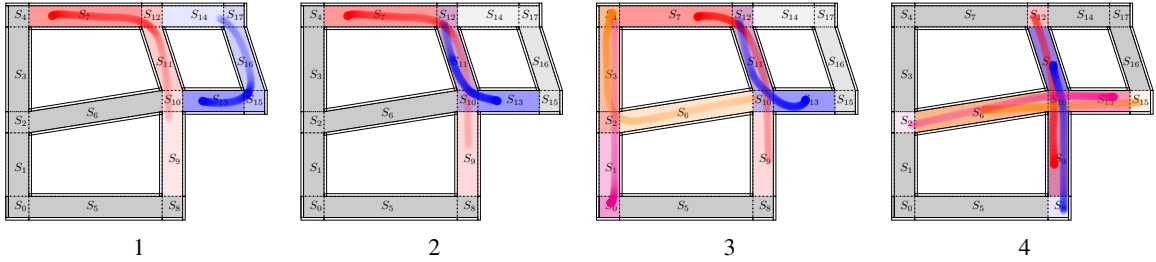

    \centering
    \begin{subfigure}[c]{0.2\linewidth}
    \resizebox{\linewidth}{!}{\input{Figures/scenarioA.tex}}
    \subcaption*{1}
    \end{subfigure}
    ~
    \begin{subfigure}[c]{0.2\linewidth}
        \resizebox{\linewidth}{!}{\input{Figures/scenarioB.tex}}
        \subcaption*{2}
    \end{subfigure}
    ~
    \begin{subfigure}[c]{0.2\linewidth}
        \resizebox{\linewidth}{!}{\input{Figures/scenarioC.tex}}
        \subcaption*{3}
    \end{subfigure}
    ~
    \begin{subfigure}[c]{0.2\linewidth}
        \resizebox{\linewidth}{!}{\input{Figures/scenarioD.tex}}
        \subcaption*{4}
    \end{subfigure}
    \caption{The tested scenarios 1-4. Agents are initialized in the darkest shade of their respective color, and are tasked with navigating through the lighter shades. Overlaid are the resulting trajectories of one of the runs of the \textbf{D} configuration.}
    \label{fig:simenvscenarios}
\end{figure*}

In order to validate the method we define four distinct scenarios. In each scenario, a number of robotic agents is tasked with completing the route they receive at the start of the simulation. The difference between scenarios is the amount of agents in the environment as well as the expected amount of interaction required to complete the pre-specified plan, i.e. the amount of overlap in the plans. For each scenario we perform multiple simulations with the starting position of each agent in the first area of the semantic plan randomly initialized at every simulation.  
The scenarios are illustrated in Figure \ref{fig:simenvscenarios}. In scenario 1, two agents are deployed within the environment. Their plans do not overlap. In scenario 2, similar to scenario 1, two agents are deployed, but their plans show partial overlap. Scenario 3 involves four agents with partially overlapping plans, but agents do not have overlapping plans with all other agents at the same time.  Finally, in scenario 4, four agents arrive at an intersection at the same time.

The performance of the algorithm in the scenarios is evaluated  in three different configurations, firstly a configuration in which agents do not cooperate (\textbf{N}ever), secondly a configuration in which agents always cooperate (\textbf{A}lways), and finally the proposed approach in which the sets of coordinating agents are formed dynamically (\textbf{D}ynamic). Results from each scenario are averaged over all simulation runs of that scenario. 
The first metric, to evaluate the performance of the proposed approach, is the task completion time of each agent. This time is computed as the time it takes for the last agent in each run to reach their final mode. We report the minimum, maximum and average over all runs and configurations of each scenario.
% Secondly, we report on the shortest distance from the static elements in the environment:
% \begin{equation}
%     d_{o} := \min_t \left[ dist(\vx_i(t), \vo) \right] - r_i \qquad \forall i \in \mathbb{A} \land \forall \vw \in \mathbb{W}
% \end{equation}
% in which $\vx_i(t)$ the state of agent $i$ at time $t$, $\vo$ the obstacle position, and $r_i$ the radius of agent $i$. Furthermore, we report the minimum distance to other agents.
% \begin{equation}
%     d_{A} := \min_t \left[ dist(\vx_i(t), \vx_j(t)) \right] - r_i - r_j \qquad \forall i,j \in \mathbb{A} \land i \neq j
% \end{equation}
% These metrics are used as a measure for the reliance on accurate localization and as a robustness against unknown disturbances. We again report on the minimum, maximum and average over all variations of each scenario.
We, furthermore, report on the time it takes for the solver to compute a solution, and the correctness and feasibility of these solutions. In the case of computation time, we report on solver time, and configuration time separately, the latter including the time required to query the graph world model and set up the MPC problem. Note that at each time step we record the maximum time over all currently active solver instances, since each solver can be executed in parallel. We again report the minimum, maximum and average over all runs of each scenario. Collisions and infeasibilities are reported on through the number of runs in which they occur.
%\todo[inline]{Explain why we take the maximum per time iteration, parallel}

\subsection{Implementation}

The algorithm described in the paper, along with an example simulation environment is available upon request. 
The algorithm is implemented in Python, with the MPC controller implemented using the do-mpc package \cite{DOMPC2017}. The resulting optimization problem is solved using CasADi \cite{CasADi2019} and the HSL\_MA27 \cite{HSL} solver in IPOPT\footnote{The max\_iter parameter of IPOPT is set to $150$ iterations.} \cite{wachter2006implementation}. The graph querying is implemented in SPARQL using the RDFLib package \cite{RDFLib}. Simulations are implemented using the built-in simulation in do-mpc, as well as in PyBullet \cite{coumans2021}. Validation results\footnote{Results were obtained on an Intel Core i5 13500 processor running Ubuntu 20.04 under Windows Subsystem for Linux 2 (WSL2)} are obtained from the built-in simulation unless explicitly mentioned.
%\todo[inline]{Scenario results are done using simple simulation because of long simulation times in PyBullet (since it is real time).}
%\subsection{Simulation Environment}
The environment in which simulations were performed is depicted in Figure \ref{fig:simenvscenarios}. The environment workspace is approximately $20 \text{ by } 23 \ [m]$ in size. The corridor width of each corridor is approximately $2.0 \ [m]$. The configuration parameters of the MPC are summarized in Table \ref{tab:param}. Velocity and acceleration ranges represent what is expected in common commercial robots. 
%During each simulation run a set of agents is deployed within the area, tasked with reaching their respective goal areas after crossing a pre-specified sequence of areas.
During the experiments, it was assumed that agents are homogeneous, more specifically, that all agents can be contained in a cylinder with a base radius of  $0.3 \ [m]$. Dynamically, agents were described by a kinematic unicycle model. %
\begin{table}[t]
    \centering
    \caption{MPC configuration parameters in simulation experiments}
    \begin{tabular}{lc|cl}
    Parameter                  & Symbol                     & value             &           \\
    \hline
    Longitudonal vel. & $[v_{min}, v_{max}]$           & {[}0.0, 1.0{]}  & {[}$m/s${]} \\
    Longitudonal acc. & $[a_{min}, a_{max}]$           & {[}-0.50, 0.50{]} & {[}$m/s^2${]} \\
    Angular vel.      & $[\omega_{min}, \omega_{max}]$ & {[}-0.50, 0.50{]} & {[}$rad/s${]} \\
    Agent radius               & $r_v$                      & 0.40               & {[}m{]}   \\
    Enlarged agent radius               & $r_{soft}$                      & 0.45               & {[}m{]}   \\
    MPC prediction time step              & $\Delta T$                 & 0.50              & {[}s{]}   \\
    MPC update frequency & $f$ & 4 & [Hz] \\
    Prediction horizon     & $N_t$                        & 25                & {[}-{]}  \\
    Cost matrices & $\vR$ & $\text{diag}(0.05,0.5)$ & [-]\\
    & $\vQ$ & 1 & [-]\\
    & $\vq$ & 10 & [-]\\
    Semantic Horizons & $N_h$, $N_{hA}$  &   2, 1 & {[}-{]} \\
    \end{tabular}
\label{tab:param}
\end{table}
\subsection{Results}
For visualization purposes, an example realization of a run of the \textbf{D} configuration of each scenario has been overlaid in Figure \ref{fig:simenvscenarios}. The remaining of this section presents the quantification of the performance metrics over 25 simulations per scenario and configuration.

\subsubsection{Completion}
For each scenario and configuration, Table \ref{tab:success} summarizes the number of non-successful task completions. It is observed that in scenario 1 all runs lead to successful task completion. 
%except for one run in both the \textbf{D} and \textbf{N} configurations in which the solver couldn't find a solution that satisfied the constraints at $t=0$.
In scenario 2 and 3, the majority of runs in the \textbf{N}ever configuration fails due to collisions. This is explained by the lack of collision constraints between agents when there is no cooperation. Two of the \textbf{D} and one of the \textbf{D} runs fail due to the solver concluding that the problem is infeasible.
From Table \ref{tab:success}, it is apparent that scenario 4 is the most challenging even in the \textbf{A} and \textbf{D} configuration. In both, at least 5 out of 25 runs fail due to minor collisions (overlapping footprints of $< 0.01 $[m]) in between sample points. All runs in the \textbf{N} configuration fail due to collisions between agents, again due to the lack of no collision constraints.
\subsubsection{Completion Time}
The task completion time for all scenarios and configurations expressed in time steps  is reported in Table \ref{tab:tcomplete}. It is apparent, from the average completion time, that the results obtained in the \textbf{A} and \textbf{D} configuration are largely comparable to each other. The \textbf{N}ever configuration does perform better, leading to shorter completion, however, this approach leads to the highest number of observed collisions, both explained by the lack of no-collision constraints in the MPC formulation. Which, in this case, allows for trajectories that do intersect each-other.
\begin{table}[]
    \centering
    \caption{Non-successful Completion of each scenario for the \textbf{A}lways, \textbf{D}ynamic and \textbf{N}ever configurations, due to infeasibilities/ collisions.}
    \begin{tabular}{cc|ccc}
         & N & \textbf{A} & \textbf{D} & \textbf{N} \\
         &   & infeas. / coll. & infeas. / coll. & infeas. / coll.\\
         \hline
    1              &    25                & 0 / 0     & 0 / 0      & 0 / 0      \\
    2              &    25                & 2 / 0     & 1 / 0      & 1 / 17        \\
    3              &    25                & 0 / 0     & 0 / 0      & 0 / 25         \\
    4              &    25                & 0 / 5     & 0 / 6      & 0 / 25            \\
\end{tabular}
\label{tab:success}
\end{table}

\begin{table}[]
    \centering
    \caption{Completion time, in time steps, of completed runs of each scenario for the \textbf{A}lways, \textbf{D}ynamic and \textbf{N}ever configurations.}
    \begin{tabular}{c|cc|c}
    $T_{task}$     & \textbf{A} & \textbf{D} & \textbf{N} \\ 
         & [min, avg, max] & [min, avg, max]& [min, avg, max]\\
         \hline
    1                                 & {[}74, 95.2, 143{]}  &  {[}78, 96.0, 138{]}     & {[}78, 96.0, 138{]}        \\
    2                                  &  {[}84, 121.1, 161{]} &  {[}84, 114.0, 142{]}     & {[}78, 102.8, 121{]}      \\
    3                                  & {[}98, 133, 167{]} & {[}97, 128.5, 155{]}       & {[}84, 124.7, 162{]}         \\
    4                                  & {[}74, 101.7, 140{]} &  {[}74, 101.4, 140{]}      & {[}81, 93.6, 116{]}                      \\
\end{tabular}
\label{tab:tcomplete}
\end{table}

\subsubsection{Computation Time}

The MPC computation times are reported in Table \ref{tab:tmpc}. The average number of sets of coordinating agents and thus MPC controllers, and the average number of agents within a these sets, are reported in Table \ref{tab:flocksize}. From Table \ref{tab:tmpc} it is observed that the MPC computation times of the \textbf{D} configuration are, on average, lower than the \textbf{A} configuration, but higher than the \textbf{N} configuration. In scenario 1, no plan overlap occurs, which leads to the \textbf{D} configuration performing, on average, the same as the \textbf{N} configuration. In scenario 4, four agents arrive at the same intersection leading to similar performance between the D and N configuration, still better than the A configuration.  

Table \ref{tab:tconfig} summarizes the recorded configuration times. When reconfiguration is necessary, i.e. when one of the agents controlled by the MPC controller crosses over into the next area, it is apparent that significant time is spent on reconfiguration of the controller. Further observations are that, on average, the \textbf{N} and \textbf{D} configurations perform better than the \textbf{A} configuration.

% \begin{figure}[t]
%     \centering
%     \includegraphics[width=1\linewidth]{Figures/t_mpc_comp.pdf}
%     \caption{The computation and configuration times for an agent in ten different runs of scenario B.0 for the \textbf{A}lways, \textbf{D}ynamic and \textbf{N}ever flock formation cases. Each color represents a separate run.}
%     \label{fig:tmpc_comparison}
% \end{figure}

\begin{table}[t]
    \centering
    \caption{MPC computation time, in milliseconds, of completed runs of each scenario for the \textbf{A}lways, \textbf{D}ynamic and \textbf{N}ever configurations. }
    \begin{tabular}{c|cc|c}
    $T_{mpc}$     & \textbf{A} & \textbf{D} & \textbf{N} \\ 
     {[}ms{]}    & [min, avg, max] & [min, avg, max] & [min, avg, max] \\
         \hline
    1                                  & {[}29, 66, 195{]}  &  {[}14, 29, 93{]}    & {[}14, 29, 89{]}        \\
    2                                  &  {[}26, 54, 345{]} &  {[}13, 34, 250{]}      & {[}14, 24, 243{]}      \\
    3                                  & {[}68, 157, 656{]} & {[}15, 50, 326{]}       & {[}17, 30, 124{]}         \\
     4                                  & {[}62, 145, 781{]} &  {[}15, 82, 768{]} & {[}14, 25, 107{]}                      \\
\end{tabular}
\label{tab:tmpc}
\end{table}
% \begin{figure}[t]
%     \centering
%     \includegraphics[width=0.6\linewidth]{Figures/Computation.pdf}
%     \caption{MPC computation time, in milliseconds, of completed runs of each scenario for the \textbf{A}lways, \textbf{D}ynamic and \textbf{N}ever configurations. Error bars signify minimum and maximum values over all runs.}
%     \label{fig:semhorizon}
% \end{figure}

\begin{table}[ht!]
    \centering
    \caption{Configuration time, in seconds, of completed runs of each scenario for the \textbf{A}lways, \textbf{D}ynamic and \textbf{N}ever configurations. }
    \begin{tabular}{c|cc|c}
    $T_{conf}$     & \textbf{A} & \textbf{D} & \textbf{N} \\ 
     {[}s{]}     & [min, avg, max] & [min, avg, max]  & [min, avg, max]  \\
         \hline
    1                                  & {[}0.18, 0.25, 0.35{]}  &  {[}0.10, 0.14, 0.28{]}    & {[}0.10, 0.14, 0.37{]}        \\
    2                                  &  {[}0.18, 0.25, 0.37{]} &  {[}0.10, 0.20, 0.37{]}      & {[}0.10, 0.15, 0.35{]}      \\
    3                                  & {[}0.36, 0.50, 0.76{]} & {[}0.10, 0.19, 0.60{]}       & {[}0.10, 0.15, 0.91{]}         \\
    4                                  & {[}0.35, 0.53, 0.80{]} &  {[}0.10, 0.35, 0.76{]} & {[}0.10, 0.15,  0.48{]}                      \\
\end{tabular}
\label{tab:tconfig}
\end{table}

\begin{table}[ht!]
    \centering
    \caption{Average number of sets of coordinating agents/average number of agents in these sets in each scenario for the \textbf{A}lways, \textbf{D}ynamic and \textbf{N}ever configurations.}
    \begin{tabular}{c|ccc}
      $\mathbb{F}$   & \textbf{A} & \textbf{D} & \textbf{N} \\
         & avg nr. / avg size & avg nr. / avg size & avg nr. / avg size
         \\ \hline
    1              & 1.0 / 2.0     & 2.0 / 1.0   &  2.0 / 1.0     \\
    2              & 1.0 / 2.0     & 1.6 / 1.4   &  2.0 / 1.0        \\
    3              & 1.0 / 4.0     & 3.2 / 1.3   &  4.0 / 1.0         \\
    4              & 1.0 / 4.0     & 3.0 / 1.9   &  4.0 / 1.0            \\
\end{tabular}
\label{tab:flocksize}
\end{table}

\section{Conclusions}

The paper presented an approach to enable a set of mobile robots to navigate in environments with an inherent structure with such knowledge encoded in a semantic graph world model. The proposed algorithms allow to dynamically (re)configure a  Model Predictive Controller (MPC) for a set of agents. The derived constraints of the MPC problem relate to environmental elements that are relevant to each of the agents described in the MPC problem as well as the inter-agent no collision constraints. We show, in simulation, that the method is applicable, and that the proposed dynamic cooperation strategy results in decreased computation times when multiple agents are in a situation in which the algorithm identifies more than one set, i.e. in situations in which the plans of some of the agents do not overlap. When all agents do have overlapping plans, the computational performance of the \lq \lq \emph{always cooperate}\rq\rq-configuration and \lq \lq \emph{dynamic cooperation}\rq\rq-configuration are similar. Furthermore, the \lq \lq \emph{dynamic cooperation}\rq\rq-configuration exhibits similar performance (other than computation time) to the \lq\lq\emph{always cooperate}\rq\rq-configuration across all tested scenarios.
% In the current approach, we formulate a centralised MPC controller to solve the local navigation problem. Future work could focus on investigating the use of a decentralised MPC approach, or on improving the currently proposed MPC formulation to increase the computational performance and its robustness. Furthermore, currently MPC configuration times take up a significant amount of time in the iterations in which configuration is necessary. Future works should focus on making the reconfiguration more computationally efficient.
Future work will look into addressing a hybrid coordination approach where discrete approaches that could handle certain situation more efficiently or robustly, like scheduling at an intersection, can be combined with MPC controllers for continuous motion control. The challenges related to the automatic configuration and instatiation of such controllers based on graph world models remain, as well as their integration with continuous layers, here addressed by the MPC controller.
% Something about incorporating more semantics, i.e. traffic rules etc.

%\todo[inline]{Possible reduction of computation time in distributing the MPC computation, however, at the cost of greater communication. Still value in keeping problems as small as possible.}

%\todo[inline]{ Explain that the flocks lend themselves for distributed implementation}

\section*{Acknowledgments}
This work has been executed as part of the Semantic navigation for Teams of Open World Robots (TOWR) research project and is part of the ICAI EAISI FAST lab

\printbibliography

\end{document}